# Insights into Performance Fitness and Error Metrics for Machine Learning

M.Z. Naser, PhD, PE
Glenn Department of Civil Engineering, Clemson University, Clemson, SC, 29634, USA
E-mail: mznaser@clemson.edu, m@mznaser.com, Website: www.mznaser.com

Amir H. Alavi, PhD
Department of Civil and Environmental Engineering, University of Pittsburgh, Pittsburgh, PA 15261, USA
E-mail: alavi@pitt.edu

**Abstract**
Machine learning (ML) is the field of training machines to achieve high level of cognition and perform human-like analysis. Since ML is a data-driven approach, it seemingly fits into our daily lives and operations as well as complex and interdisciplinary fields. With the rise of commercial, open-source and user-catered ML tools, a key question often arises whenever ML is applied to explore a phenomenon or a scenario: *what constitutes a good ML model?* Keeping in mind that a proper answer to this question depends on a variety of factors, this work presumes that *a good ML model is one that optimally performs and best describes the phenomenon on hand*. From this perspective, identifying proper assessment metrics to evaluate performance of ML models is not only necessary but is also warranted. As such, this paper examines a number of the most commonly-used performance fitness and error metrics for regression and classification algorithms, with emphasis on engineering applications.

## 1. Introduction

Learning is the process of seeking knowledge [1]. We, as humans, can learn from our daily interactions and experiences because we have the ability to communicate, reason and understand. With the rapid technological advancement in computer sciences, computational intelligence has led to the development of modern cognitive and evaluation tools [2,3]. One such tool is machine learning (ML) which is often described as a set of methods that, when applied, can allow machines to learn/understand meaningful patterns from data repositories; while maintaining minimal human interaction [4]. More specifically, a *"computer program is said to learn from experience E with respect to some class of tasks T and performance measure P, if its performance at tasks in T, as measured by P, improves with experience E"* [5]. In other words, ML trains machines to understand real-world applications, use this knowledge to carry out pre-identified tasks with a goal of optimizing and improving the machines' performance with time and new knowledge. A closer look in the definition of ML infers that computers do not learn by reasoning, but rather by algorithms.

From the perspective of this work, traditional statistical regression techniques are often used to carry out behavioral modeling purposes wherein such techniques can suffer from large uncertainties, need for idealization of complex processes, approximation, and averaging widely varying prototype conditions. Furthermore, a statistical regression analysis often assumes linear, or in some cases nonlinear, relationships between the output and the predictor variables and these assumptions do not always hold true. On the other hand, ML methods adaptively learn from experiences and extract various discriminators. One of the major advantages of ML approaches, over the traditional statistical techniques, is their ability to derive a relationship(s) between inputs and outputs without assuming prior forms or existing relationships. In other words, ML approaches are not confined into one particular space that requires the availability of physical representation, but rather goes beyond that to explore hidden relations in data patterns [6–11].

While ML was initially developed for computer sciences, it is now an integral part of various fields including, energy/mechanical engineering [6–9], social sciences [10,11], space applications [12,13], among others [14–19]. Due to the availability of high-computationally powered machines and ease-of-access to data (thanks in part to the rise of Internet-of-Things and data-driven-applications), the utilization of ML into civil engineering, in general, and materials science, engineering in particular, has been duly noted in recent years [20–25].

An integral part of the wide spread of integrating ML into new research areas is due to the availability of user-friendly and easy-to-use software packages that simplifies the process of ML by utilizing pre-defined algorithms and training/validation procedure [26–30]. The availability of such tools, while facilitate ML analysis and provides new opportunities for researchers often unfamiliar with the ML fundamentals with means to easily carry out such analysis, could still be mis-used by providing a false sense of analysis interpretation [31]. Another concern of utilizing user-ready approaches to carry out ML analysis lies in the need for compiling proper observations (i.e. datapoints). In some classical fields (say material sciences, earthquake or fire engineering) where there is limited number of observations due to expensive tests, or need for specialized instrumentation/facilities [32], then the use of ML may lead to biased outcome – especially when combined with lack of expertise on ML [33,34].

An examination of open literature rises few questions: 1) are we developing accurate ML models? 2) are such models useful to our fields? 3) are we properly validating ML models? And 4) how to confidently answer "yes" to the aforementioned questions?

A distinction should be drawn in which we need to acknowledge that, we often apply existing ML algorithms to our problems, rather than developing new algorithms. This acknowledgement goes hand in hand with that similar to applying other numerical tools such as finite element method, to investigate the response of materials and structures (say concrete beams) under harsh environments (i.e. fire conditions) [35,36]. From this perspective, we use an existing tool, say a finite element (FE) software (ANSYS [37], ABAQUS [38] etc.), to investigate how failure mechanism occurs in a concrete beam under fire. The accuracy of this FE model is often established through a validation procedure in which a comparison of predictions from the FE model (say temperature rise in steel rebars or mid-span deflection during fire, or in some cases point in time when the beam fails) is plotted against that measured in an actual fire test. If the comparison deemed well, then the FE model is said to be valid and hence can be used to explore the effect of key response parameters (i.e. magnitude of loading, strength of concrete, intensity of fire etc.). From this perspective, the validity of an FE model is established if the variation between predicted results and measured observations is between 5-15%* [39].

*One should note that the validation of an FE model is also governed by satisfying convergence criteria input in the FE software. More on this can be found elsewhere [37,38].

Unlike the use of FE simulation, ML is often used in two domains: 1) to show the applicability of ML to understand a phenomenon [40,41], and 2) to identify hidden patterns governing a phenomenon [33,42]. In the first domain, ML is primarily used to show that an ML algorithm can replicate a phenomenon – or in other words to validate the applicability of that particular ML algorithm to a material science problem (i.e. can deep learning be applied to predict the compressive strength of concrete given that information regarding the components in a concrete mix is available?). While works in this domain showcases the diversity of ML, these also provide an additional validation platform/case studies to already well-established algorithms. The contribution of such works to our knowledge base is to be thanked and acknowledged.

The second domain is where ML shines and can be proven as a powerful ally to researchers. This is because, ML strives on data and is designed to explore hidden features and patterns. The integration of these two items has not been thoroughly applied into our fields and if applied properly cannot only open new opportunities, but also revolutionize our perspective into our fields. Unfortunately, the open literature continues to lack works in this domain and hence such works are to be encouraged.

Whether ML is used in the first or second domain, ML models need to be rigorously assessed [43,44]. This is a critical key to ensure: 1) the validity of the developed ML model in understanding a complex phenomenon given a limited set of data points, and 2) proper extension of the same models towards new/future datasets. Traditionally, the adequacy of ML models is often established through performance fitness and error metrics (PFEMs). Performance and error measures are vital elements in the process of evaluating ML models/frameworks. These are defined as logical and/or mathematical constructs intended to measure the closeness of actual observations to that expected (or predicted). In other words, PFEMs are used to establish an understanding of how predictions from a model compares to real (or measured) observations. Such metrics often relates the variation between predicted and measured observations in terms of errors [45–47].

Diverse sets of performance metrics have been noted in the open literature i.e. correlation coefficient ($R$), root mean squared error (RMSE), etc. In practice, one, a multiple or a combination of metrics are used to examine the adequacy of a particular ML model. However, there does not seem to be a systematic view into which scenarios certain metrics are preferable to use. In order to bridge this knowledge gap, this work compiles the commonly-used PFEMs and highlights their use in evaluating performance of regression and classification ML models.

## 2. Performance Fitness and Error Metrics

This section presents the most widely-used PFEMS and highlights fundamentals, recommendations and limitations associated with their use in assessing ML models[†]. In this work, PFEMs are grouped under two categories; traditional and modern. In this section, these reoccurring terms are used; $A$: actual measurements, $P$: predictions, $n$: number of data points.

[†] It should be noted that other works have used a different classification for PFEMs [2]. Botchkarev [2] went even further to survey the most preferred metrics reported by researchers during the 1980-2007 era and also explored multiplication and addition point distance methods.

*2.1 Regression*

Regression ML methods deal with predicting a target value using independent variables. Some of these methods include: artificial neural networks, genetic programing, etc. PFEMs grouped herein belong to a group of metrics that are based on methods to calculate point distance primarily using subtraction or division operations. These metrics contain fundamental operations either *A-P* or *P/A* and can be supplemented with absoluteness or squareness. These are the most widely-used metrics in literature. The simplest form of common PFEMs results from subtracting a predicted value from its corresponding actual/observed value. This is often straightforward, easy to interpret and most of all yield the magnitude of error (or difference) in the same units as those measured and predicted and can indicate if the model overestimates or underestimates observations (by analyzing the sign of the reminder). One should remember that an issue could arise where due to opposite between predictions and observations i.e. canceling positive and negative error. In this scenario, a zero error could be calculated, indicating false accuracy.

This can be avoided by using an absolute error (i.e. |*A-P*|) which only yields non-negative values. Analogous to traditional error, absolute error also maintains the same units of predictions (and observations); and hence is easily relatable. However, due to its nature, the bias in absolute errors cannot be determined.

Similar to the same concept of absolute error, squared error also mitigates mutual cancellation of errors. This metric can be continuously differentiable and thus facilitates optimization. However, this metric emphasizes relatively large errors (as opposed to small errors), unlike absolute error, and could be susceptible to outliners. The fact that the units of squared error is squared lead to unconventional units for error (i.e. squared days); which are not intuitive. Other metrics may also include logarithmic quotient error (i.e. *ln(P/A)*) as well as absolute logarithmic quotient error (i.e. |*ln(P/A)*|). Table 1 lists other commonly used metrics, together with some of their limitations and shortcomings as identified by surveyed studies.

Table 1 List of commonly used PFEMs for ML regression models as collected from open literature

| No. | Metric | Definition | Formula | Remarks |
|---|---|---|---|---|
| 1 | Error (E) | The amount by which an observation differs from its actual value. | $E = A - P$ | • Intuitive<br>• Easy to apply |
| 2 | Mean error (ME) | The average of all errors in a set. | $ME = \frac{\sum_{i=1}^{n} E_i}{n}$ | • May not be helpful in cases where positive and negative predictions cancel each other out. |
| 3 | Mean Normalized Bias (MNB) | Associated with observation-based minimum threshold. | $MNB = \frac{\sum_{i=1}^{n} E_i/A_i}{n}$ | • Biased towards overestimations. |
| 4 | Mean Percentage Error (MPE) | Computed average of percentage errors. | $MPE = \frac{\sum_{i=1}^{n} E_i/A_i}{n/100}$ | • Undefined whenever a single actual value is zero. |
| 5 | Mean Absolute Error (MAE)* | Measures the difference between two continuous variables. | $MAE = \frac{\sum_{i=1}^{n} \lvert E_i \rvert}{n}$ | • Uses a similar scale to input data [48].<br>• Can be used to compare series of different scales. |
| 6 | Mean Absolute Percentage Error (MAPE)* | Measures the extent of error in percentage terms. | $MAPE = \frac{100}{n} \sum_{i=1}^{n} \lvert E_i \rvert / \lvert A_i \rvert$ | • Commonly-used as a loss function [49]<br>• Cannot be used if there are actual zero values.<br>• Percentage error cannot exceed 1.0 for small predictions.<br>• There is no upper limit to percentage error in predictions that are too high.<br>• Non-symmetrical (adversely affected if a predicted value is larger or smaller than the corresponding actual value) [49]. |
| 7 | Relative Absolute Error (RAE) | Expressed as a ratio comparing the mean error to errors produced by a trivial model. | $RAE = \sum_{i=1}^{n} \lvert E_i \rvert / \lvert A_i - A_{mean} \rvert$ | • $E_i$ ranges from zero (being ideal) to infinity. |
| 8 | Mean Absolute Relative Error (MARE) | Measures the average ratio of absolute error to random error. | $MARE = \frac{1}{n} \sum_{i=1}^{n} \lvert E_i \rvert / \lvert A_i \rvert$ | • Sensitive to outliers (especially of low values).<br>• Division by zero may occur (if actuals contain zeros). |
| 9 | Mean Relative Absolute Error (MRAE) | Ratio of accumulation of errors to cumulative error of random error. | $MRAE = \frac{\sum_{i=1}^{n} \lvert E_i \rvert / \lvert A_i - A_{mean} \rvert}{n}$ | • For a perfect fit, the numerator equals to zero [50]. |
| 10 | Geometric Mean Absolute Error (GMAE)* | Defined as the n-th root of the product of error values. | $GMAE = \sqrt[n]{\prod_{i=1}^{n} \lvert E_i \rvert}$ | • GMAE is more appropriate for averaging relative quantities as opposed to arithmetic mean [51].<br>• This metric can be dominated by large outliers and minor errors (i.e. close to zero). |
| 11 | Fractional Absolute Error (FAE) | Evaluates the absolute fractional error. | $FAE = \frac{1}{n} \sum_{i=1}^{n} \frac{2 \times \lvert E_i \rvert}{\lvert A_i \rvert + \lvert P_i \rvert}$ | - |
| 12 | Mean Squared Error (MSE) | Measures the average of the squares of the errors. | $MSE = \frac{\sum_{i=1}^{n} {E_i}^2}{n}$ | • Scale dependent [52].<br>• Values closer to zero present adequate state<br>• Heavily weights outliers.<br>• Highly dependent on fraction of data used (low reliability) [53]. |
| 13 | Root Mean Squared Error (RMSE) | Root square of average squared error. | $RMSE = \sqrt{\frac{\sum_{i=1}^{n} {E_i}^2}{n}}$ | • Scale dependent.<br>• A lower value for RMSE is favorable.<br>• Sensitive to outliers.<br>• Highly dependent on fraction of data used (low reliability) [53]. |
| 14 | Sum of Squared Error (SSE) | Sums the squared differences between each observation and its mean. | $SSE = \sum_{i=1}^{n} {E_i}^2$ | • A small SSE indicates a tight fit [54]. |
| 15 | Relative Squared Error (RSE) | Normalizes total squared error by dividing by the total squared error. | $RSE = \sum_{i=1}^{n} {E_i}^2 / (A_i - A_{mean})^2$ | • A perfect fit is achieved when the numerator equals to zero [50]. |
| 16 | Root Relative Squared Error (RRSE) | Evaluates the root relative squared error between two vectors. | $RRSE = \sqrt{\sum_{i=1}^{n} {E_i}^2/(A_i - A_{mean})^2}$ | • Ranges between zero and 1, with zero being ideal [50]. |
| 17 | Geometric Root Mean Squared Error (GRMSE) | Evaluates the geometric root squared errors. | $GRMSE = \sqrt[2n]{\prod_{i=1}^{n} {E_i}^2}$ | • Scale dependent.<br>• Less sensitive to outliners than RMSE [52]. |
| 18 | Mean Square Percentage Error (MSPE)* | Evaluates the mean of square percentage errors. | $MSPE = \frac{\sum_{i=1}^{n}(\lvert E_i\rvert/\lvert A_i\rvert)^2}{n/100}$ | • Non-symmetrical [49]. |
| 19 | Root Mean Square Percentage Error (RMSPE)* | Evaluates the mean of squared errors in percentages. | $RMSPE = \sqrt{\frac{\sum_{i=1}^{n}(\lvert E_i\rvert/\lvert A_i\rvert)^2}{n/100}}$ | • Scale independent.<br>• Can be used to compare predictions from different datasets.<br>• Non-symmetrical [49]. |
| 20 | Normalized Root Mean Squared Error (NRMSE)** | Normalizes the root mean squared error. | $NRMSE = \frac{\sqrt{\frac{\sum_{i=1}^{n} {E_i}^2}{n}}}{A_{mean}}$ | • Can be used to compare predictions from different datasets [55]. |
| 21 | Normalized Mean Squared Error (NMSE) | Estimates the overall deviations between measured values and predictions. | $NMSE = \frac{\frac{\sum_{i=1}^{n} {E_i}^2}{n}}{variance^2}$<br>$variance = \frac{\sum(x_i - mean)^2}{n-1}$ | • Biased towards over-predictions [56]. |
| 22 | Coefficient of Determinatio n ($R^2$) | The square of correlation. | $R^2 = 1 - \sum_{i=1}^{n}(P_i - A_i)^2 / \sum_{i=1}^{n}(A_i - A_{mean})^2$ | • $R^2$ values close to 1.0 indicate strong correlation.<br>• Can be used in predicting material properties. |
| 23 | Correlation coefficient (R) | Measures the strength of association between variables. | $R = \frac{\sum_{i=1}^{n}(A_i - \overline{A}_i)(P_i - \overline{P}_i)}{\sum_{i=1}^{n}(A_i - \overline{A}_i)^2 \sum_{i=1}^{n}(P_i - \overline{P}_i)}$ | • R>0.8 implies strong correlation [57].<br>• Does not change by equal scaling.<br>• Can be used in predicting |
| | | | | material properties. |
| 24 | Mean Absolute Scaled Error (MASE) | Mean absolute errors divided by the mean absolute error. | $\frac{\sum_{i=1}^{n}\frac{E_i}{A_i}}{n/100}/(\frac{1}{n}-1)\sum_{i=1}^{n}\lvert A_i - A_{i-1}\rvert$ | • Scale independent.<br>• Stable near zero [52]. |
| 25 | Golbraikh and Tropsha's [58] criterion | - | *At least one slope of regression lines (k or k') between the regressions of actual ($A_i$) against predicted output ($P_i$) or $P_i$ against $A_i$ through the origin, i.e. $A_i = k \times P_i$ and $t_i = k' A_i$, respectively.*<br>$k = \frac{\sum_{i=1}^{n}(A_i \times P_i)}{{A_i}^2}$<br>$k' = \frac{\sum_{i=1}^{n}(A_i \times P_i)}{{A_i}^2}$<br>$m = \frac{R^2 - {R_o}^2}{R^2}$<br>$n = \frac{R^2 - {R_o'}^2}{R^2}$ | • $k$ and $k'$ need to be close to 1 or at least within the range of 0.85 and 1.15.<br>• $m$ and $n$ are performance indexes and their absolute value should be lower than 0.1. |
| 26 | QSAR model by Roy and Roy [59] | - | $R_m = R^2 \times (1 - \sqrt{\lvert R^2 - {R_o}^2\rvert})$<br>*where,*<br>${R_o}^2 = \frac{\sum_{i=1}^{n}(P_i - A_i^o)^2}{\sum_{i=1}^{n}(P_i - P_{mean})^2}, A_i^o = k \times P_i$<br>${R_o}^2 = \frac{\sum_{i=1}^{n}(A_i - P_i^o)^2}{\sum_{i=1}^{n}(A_i - A_{mean})^2}, P_i^o = k \times A_i$ | • $R_m$ is an external predictability indicator. $R_m > 0.5$ implies a good fit. |
| 27 | Frank and Todeschini [60] | - | *Recommend maintaining a ratio of 3-5 between the number of observations and input parameters.* | - |
| 28 | Objective function by Gandomi et al. [61] | A multi-criteria metric. | *Function*<br>$= (\frac{No._{Training} - No._{Validation}}{No._{Training} + No._{Validation}})\frac{RMSE}{}$<br>$+ \frac{2No._{Validation}}{No._{Training} + No._{Validation}}\frac{RMSE_{Va}}{}$<br>*where, $No._{Training}$ and $No._{Validation}$ are the number of training and validation data, respectively.* | • This function needs to be minimized to yield highest fitness.<br>• Can be used in predicting material properties. |
| 29 | Reference index (RI) by Cheng et al. [62] | A multi-criteria metric that uniformly accounts for RMSE, MAE and MAPE. | $RI = \frac{RMSE + MAE + MAPE}{3}$ | • Each fitness metric is normalized to achieve the best performance. |

*has a median derivative

**can be normalized by standard deviation of actual observations

***The reader is encouraged to review the cited references for full details on specific metrics.

Most of the works conducted so far in the areas of engineering applications only utilized a few of the above PFEMs [20,33,61–82][83]. The bulk of the reviewed works continue to incorporate traditional metrics such as *R*, $R^2$, *MAE*, *MAPE*, and *RMSE* as primary indicators of adequacy of the regression-based ML models. This seems to stem from our familiarity with these indicators, as opposed to others; such as Golbraikh and Tropsha's [58] criterion, QSAR model by Roy and Roy [59], Frank and Todeschini [60], and specifically designed objective functions, often used in the realms of other fields and data sciences. It should be noted that out of the reviewed studies, the works of Gandomi et al. [61,79,80], Golafshani and Behnood [84] as well as Cheng et al. [62] applied a multi-criteria verification process which incorporated the use of traditional as well as modern PFEMs. Utilizing multi-criteria is not only beneficial to ensure validity of a particular ML model but is also recommended to overcome some of the identified limitations of traditional metrics in Table 1 and hence should be encouraged.

### *2.2 Classification*

In ML, classification refers to categorizing data into distinct classes. This is a supervised learning approach where machines learn to classify observations into binary or multi-classes. Binary classes are those with two labels (i.e. positive vs. negative etc.) and multi-classes are those having more than two labels (i.e. types of concrete e.g. normal strength, high strength, high performance etc.). Classification algorithms may include logistic regression, k-nearest neighbors, support vector machines, etc. [85,86].

The performance of classifiers is often listed in a confusion matrix. This matrix contains statistics about actual and predicted classifications and lays the fundamental foundations necessary to understand accuracy measurements for a specific classifier. Each column in this matrix signifies predicted instances, while each row represents actual instances. This matrix was identified to be the "go-to" metric used in studies examining materials science and engineering problems [22,87–90]. However, there are other PFEMs that can be used to evaluate classification models, and these, along with others, are listed in Table 2. Similar to Table 1, Table 2 also lists some of the remarks and limitations pointed out by surveyed works. In this table, *P (denotes number of real positives), N (denotes number of real negatives), TP (denotes true positives), TN (denotes true negatives), FP (denotes false positives), and FN (denotes false negatives).*

Table 2 List of the commonly-used PFEMs for ML classification models as collected from open literature

| No. | Metric | Definition | Formula | Remarks |
|---|---|---|---|---|
| 1 | True Positive Rate (TPR) or Sensitivity or Recall | Measures the proportion of actual positives that are correctly identified as positives. | $TPR = \frac{TP}{P} = \frac{TP}{TP + FN} = 1 - FNR$ | • Describes the proportion of actual positives that are correctly identified.<br>• Does not account for indeterminate results. |
| 2 | True Negative Rate TNR or | Measures the proportion of actual negatives | $TNR = \frac{TN}{N} = \frac{TN}{TN + FP} = 1 - FPR$ | • Describes the proportion of actual |
| | Specificity or selectivity | that are correctly identified negatives. | | negatives that are correctly identified. |
| 3 | Positive Predictive Value (PPV) or Precision | The proportions of positive observations that are true positives. | $PPV = \frac{TP}{TP + FP} = 1 - FDR$ | • Has an ideal value of 1 and the worst value of zero. |
| 4 | Negative Predictive Value (NPV) | The proportions of negative observations that are true positives. | $NPV = \frac{TN}{TN + FN} = 1 - FOR$ | • Has an ideal value of 1 and the worst value of zero. |
| 5 | False Positive Rate (FPR) | Measures the proportion of positive cases in that are correctly identified as positives. | $FPR = \frac{FP}{N} = \frac{FP}{FP + TN} = 1 - TNR$ | • Describes proportion of negative cases incorrectly identified as positive cases. |
| 6 | False Discovery Rate (FDR) | Expected proportion of false observations. | $FDR = \frac{FP}{FP + TP} = 1 - PPV$ | • Describes proportion of the individuals with a positive test result for which the true condition is negative. |
| 7 | False Omission Rate (FOR) | Measures the proportion of false negatives that are incorrectly rejected. | $FDR = \frac{FN}{FN + TPN} = 1 - NPV$ | • Describes proportion of the individuals with a negative test result for which the true condition is positive. |
| 8 | Positive likelihood ratio (LR+) | Evaluates the change in the odds of having a diagnosis with a positive test. | $LR+= \frac{TPR}{FPR}$ | • Measures the ratio of TPR (sensitivity) to the FPR (1 – specificity).<br>• Presents the likelihood ratio for increasing certainty about a positive diagnosis. |
| 9 | Negative likelihood ratio (LR-) | Evaluates the change in the odds of having a diagnosis with a negative test. | $LR-= \frac{FNR}{TNR}$ | • Describes the ratio of FNR to TNR (specificity). |
| 10 | Diagnostic odds ratio (DOR) | Measures the effectiveness of a (diagnostic) test. | $DOR = \frac{LR+}{LR-} = \frac{TP/FP}{FN/TN}$ | • Often used in binary classification. |
| 11 | Accuracy (ACC) | Evaluates the ratio of number of correct predictions to the total number of samples. | $ACC = \frac{TP + TN}{P + N} = \frac{TP + TN}{TP + TN + FP + FN}$ | • Presents performance at a single class threshold only.<br>• Assumes equal cost for errors [88]. |
| 12 | $F_1$ score | Harmonic mean of the precision and recall. | $F_1 = \frac{2PPV \times TPR}{PPV + TPR} = \frac{2TP}{2TP + FP + FN}$ | • Describes the harmonic mean of precision and sensitivity.<br>• Focuses on one class only.<br>• Biased to the majority class [91]. |
| 13 | Matthews Correlation Coefficient (MCC) | Measures the quality of binary classifications analysis. | $MCC = \frac{TP \times TN - FP \times FN}{\sqrt{(TP + FP)(TP + FN)(TN + FP)}}$ | • Measures the quality of binary and multi-class classifications.<br>• Can be used in classes with different sizes.<br>• When MCC equals +1 → perfect prediction, → 0 equivalent to a random prediction and → −1 false prediction.<br>• Considered as a balanced measures as it involves values of all the four quardants of a confusion matrix [92]. |
| 14 | Bookmaker Informedness (BM) or Youden's J statistic | Evaluates the discriminative power of the test [93]. | $BM = TPR + TNR - 1$ | • Describes the probability of an informed decision (vs. a random guess).<br>• Has a range between zero and 1 (being ideal).<br>• Considers both real positives and real negatives.<br>• Takes into account all predictions [94].<br>• Counterpart of recall.<br>• It is also suitable with imbalanced data.<br>• It does not change concerning the differences between the sensitivity and specificity [93]. |
| 15 | Markedness (MK) | Measures trustworthiness of positive and negative predictions. | $MK = PPV + NPV - 1$ | • Measures trustworthiness of positive and negative predictions by a model [95].<br>• Considers both predicted positives and predicted negatives. |
| | | | | • Counterpart of precision.<br>• Specifies the probability that a condition is marked by the predictor (as opposed to luck/chance) [96]<br>• Sensitive to data changes (not suitable for imbalanced data) [93]. |
| 16 | Average Class Accuracy (ACA) | Measures the average accuracy of predictions in a class. | $ACA = W\left(\frac{TP}{TP+FP}\right) + (1 - W)\left(\frac{TN}{TN+FP}\right)$ $where\ \ 0 < W < 1$ | • Used with unbalanced data.<br>• Choosing a good weighting factor a priori [91].<br>• When $W > 0.5$, minority class accuracy contributes more than majority class.<br>• Presents performance at a single class threshold. |
| 17 | Receiver Operating Characteristic (ROC) | Plots the diagnostic ability of a binary classifier system as its discrimination threshold is varied. | *The ROC curve is plotted such that TPR is on vertical axis and FPR is on the horizontal axis (the line TPR = FPR represents a random guess of a specific class)* [97]. | • Characterizes tradeoff between hit rate and false alarm rate.<br>• Designates the relationship between sensitivity and specificity [98].<br>• Takes a value between zero and 1 to relate the probability distribution to a single state [99].<br>• A threshold of zero ensures highest sensitivity and 1 ensures best specificity.<br>• Can be used to estimate cost ratio (slope of line tangent to ROC curve).<br>• Should be used in datasets with roughly equal numbers of observations for each class [100,101]. |
| 18 | Area under the ROC curve (AUC) | Measures the two-dimensional area underneath the entire ROC curve. | $AUC = \sum_{i=1}^{N-1} \frac{1}{2}(FP_{i+1} - FP_i)(TP_{i+1} - TP_i)$ | • Not dependent on a single class threshold. |
| | | | *or* $AUC = \frac{1}{2} w\,(h + h'),$ *where, w = width, and h and h' = heights of the sides of a trapezoid histogram* | • Associated with increased training times. |
| 19 | Precision-Recall curve | Plots the tradeoff between precision and recall for different thresholds. | *Plots precision (in the vertical axis) and the recall (in the horizontal axis) for different thresholds.* | • Applicable in cases of moderate to large class imbalance [100].<br>• Used in binary classification. |
| 20 | Log Loss Error (LLE) | Measures the where the prediction input is a probability value. | $LLE = -\sum_{c=1}^{M} A_i logP,$ *where, M: number of classes, c: class label, y: binary indicator (0 or 1) if c is the correct classification for a given observation.* | • Measures the uncertainty of the probabilities by comparing predictions to the true labels.<br>• Penalizes for being too confident in wrong prediction.<br>• Has probability between zero and 1.<br>• A log loss of zero indicates a perfect model. |
| 21 | Hinge Loss Error (HLE) | - | $HLE = max(0,1 - q \cdot y)$ *where, q= ±1 and y: classifier score* | • Linearly penalize incorrect predictions.<br>• Primarily used in support vector machine. |
| 22 | Wilcoxon–Mann–Whitney (WMW) test [91] | - | $WMW = \frac{\sum_{i \in Minor\ class} \sum_{i \in Major\ class} I_{wmw}(P_i}{\lvert Minor\ class \rvert \times \lvert Major\ class \rvert}$ *where, $P_i$ and $P_j$: outputs when evaluated on an example from the minority and majority classes, respectively* | • Used in scenarios with unbalanced data.<br>• The indicator function $I_{wmw}$ returns 1 if $P_i > P_j$ and $P_i \geq 0$ or 0 if otherwise. |
| 23 | Fitness Function *Amse* (FFA) [91] | Measures pattern difference between input and output. | $FFA = \frac{1}{K} \sum_{c=1}^{K} \left(1 - \frac{\sum_{i=1}^{N_c} (1 - sig(P_{ci}) - T_c)}{N_c \times 2}\right)^2,$ $sig(x) = \frac{2}{1 + e^{-x}} + 1$ *where, $P_{ci}$: output of a classifier evaluated on the ith example, $N_c$: number of examples, K: number of classes, $T_c$ : target values (equals to -0.5 and 0.5 for majority and minority classes, respectively)* | • Used in scenarios with unbalanced data.<br>• Appropriate for genetic programing.<br>• Needs to be scaled to a range of [-1, 1] and hence the need for sigmoid function.<br>• FFA = 1 presents an ideal scenario. |
| 24 | Fitness Function *Incr* (FFI) [91] | - | $Incr = \frac{1}{K}\sum_{c=1}^{K}\left(\frac{\sum_{j=1}^{M_c}\left[I_{zt}\left(j, D_{cj}, c\right). \sum_{i=1}^{N_c} Eq(\ldots}{\frac{1}{2}N_c(N_c+1)}\right.$<br>$I_{zt}(r,k,c) = \begin{cases} r, & if\ k \geq 0\ and\ c \in \ldots \\ & or\ if\ k < 0\ and\ c \ldots \\ 0, & other\ldots \end{cases}$<br>$Eq(p,q) = \begin{cases} 1, \\ 0, otherwise \end{cases}$ | • Used in scenarios with unbalanced data.<br>• Assigns incremental rewards to predictions that fall further away from the class boundary.<br>• Appropriate for genetic programming.<br>• Ranges [0, 1] (zero being worst fitness). |
| 25 | Fitness Function Correlation (FFC) | - | $FFC = \frac{1}{K}\left(r + I_{zt}(1, \mu_{minor}, \mu_{major}\right),$<br>$r = \sqrt{\frac{\sum_{c=1}^{K} N_c(\mu_c - \bar{\mu})^2}{\sum_{c=1}^{K}\sum_{i=1}^{N_c}(P_{ci} - \bar{\mu})^2}}$<br>$\mu_c = \frac{\sum_{i=1}^{N_c} P_{ci}}{N_c}, \bar{\mu} = \frac{\sum_{c=1}^{K} N_c\mu_c}{\sum_{c=1}^{K} N_c}.$<br>*where, r: correlation ratio, $\mu_{minor}$ and $\mu_{major}$: mean for minor and major classes, respectively* | • Used in scenarios with unbalanced data. |
| 26 | Fitness Function Distribution (FFD) | Measures the distance between class distributions as a function of class separability. | $FFD = \frac{\left\lvert\mu_{min} - \mu_{maj}\right\rvert}{\sigma_{min} + \sigma_{maj}} \times I_{zt}\left(2, \mu_{min}, \mu_{maj}\right)$<br>$\mu_c = \frac{\sum_{i=1}^{N_c} P_{ci}}{N_c}, \sigma_c = \sqrt{\frac{1}{N_c}\sum_{i=1}^{N_c}(P_{ci} - \mu_c)^2}.$<br>*where, $\mu_c$ and $\sigma_c$: mean and standard deviation of the class distribution, respectively,* | • Used in scenarios with unbalanced data.<br>• Treats predictions as independent distributions.<br>• Measures separability (i.e. distance between class distributions) [102] – high separability (no overlap) and this distance turns large (go to $+\infty$).<br>• Uses $I_{zt}$ to enforce zero class threshold. |
| 27 | Canberra Metric (CM) | Measures the distance between pairs of points in a vector space. | $CM = \sum_{i=1}^{n}\frac{\lvert E_i\rvert}{A_i + P_i}$ | - |
| 28 | Wave Hedges Distance (WHD) | - | $WHD = \sum_{i=1}^{n}\frac{\lvert E_i\rvert}{max\ (A_i, P_i)}$ | • Normalizes the difference of each pair of coefficients with its maximum [103–105]. |
| 29 | Lift [106] | Measures the performance of a model at predicting or classifying cases. | $LIFT = \frac{\%of\ true\ positives\ above\ the\ th\ldots}{\%of\ dataset\ above\ the\ thres\ldots}$ | • Measures betterness of a classifier than a baseline classifier that randomly predicts positives. |
| | | | | • Threshold is set as a static fraction of the positive dataset.<br>• Lift and Accuracy do not always correlate well. |
| 30 | Mean Cross Entropy (MXE) | Measures the performance of a model where the output is a probability between zero and one. | $MXE = -\frac{1}{N}\sum True \times ln(Predicted) + (1 - True) \times ln(1 - Predicted)$<br><br>*(The assumptions are that Predicted ∈ [0, 1] and True ∈ {0, 1})* | • Minimizing MXE gives the maximum likelihood [94]. |
| 31 | Probability Calibration (CAL) | - | *1. Order cases 1-100 by their predicted in the same bin.*<br>*2. Evaluate the percentage of true positives.*<br>*3. Calculate the mean prediction for true positives.*<br>*4. Calculate the mean prediction calibration error for this bin (using the absolute value of the difference between the observed frequency and the mean).*<br>*5. Repeat steps 1-4 for cases 2-101, 3-102, etc.*<br>*6. CAL is calculated as the mean of these binned calibration errors* [94]. | • Lengthy procedure. |
| 32 | Precision-recall break-even point | Point at which the precision-recall-curve intersects the bisecting line. | *Precision = Recall* | • Defines the point when precision and recall are equal. |
| 33 | Average precision (AP) | Combines recall and precision for ranking. | $AP = \sum_{n}(Recall_n - Recall_{n-1})Percision_n$ | • Describes the weighted mean of precision in each threshold with the increase in recall from the previous threshold used. |
| 34 | Balanced accuracy [107] | Calculates the average of the correctly identified proportion of individual classes. | *Defined as the average of recall obtained on each class.* | • Used in binary and multiclass classification problems.<br>• Accommodates imbalanced datasets. |
| 35 | Brier score (BS) | Measures the accuracy of probabilistic-based predictions. | $BS = \frac{1}{N}\sum_{i=1}^{N}(f_i - A_i)^2$<br><br>*in which $f_i$ is the probability that was forecast, $A_i$ the actual outcome of the event at instance i* | • Measures the mean squared difference between the predicted probability and the actual outcome.<br>• Takes on a value between zero and 1 |
| | | | | (the lower the score is, the better the predictions).<br>• Composed of refinement loss and calibration loss.<br>• Appropriate for binary and categorical outcomes.<br>• Inappropriate for ordinal variables. |
| 36 | Cohen's kappa (CK) [108] | Measures interrater (agreement) reliability. | $\kappa = (p_o - p_e)/(1 - p_e)$<br>*where, $p_o$: empirical probability of agreement on the label assigned to any sample, $p_e$: expected agreement when both annotators assign labels randomly and this is estimated using a per-annotator empirical prior over the class labels.* | • Measures inter-annotator agreement.<br>• Expresses the level of agreement between two annotators [109].<br>• Ranges between -1 and 1. The maximum value means complete agreement. |
| 37 | Hamming loss (HL) | Fraction of the wrongly identified labels. | $HL = \frac{1}{m}\sum_{i=1}^{m} 1_{\widehat{P_i \neq A_i}}$ | • Describes fraction of labels that are incorrectly predicted.<br>• Optimal value is zero [110]. |
| 38 | Fitness (T) [111] | - | $Fitness(T) = Q(T) + \alpha * R(T) + \beta * Cost(T)$<br><br>*where, Q(T): accuracy, R(T): sum of R($T_i$) in all multi-tests of the T tree, Cost(T): sum of the costs of attributes constituting multi-tests. The default parameters values are: α=1.0 and β=−0.5,*<br>$R(T_i) = \frac{\lvert X_i \rvert}{\lvert X \rvert} * \sum_{j=1}^{\lvert mt_i \rvert - 1} r_{ij}$<br>*where, X: learning set, $X_i$: instances in i-th node, and $\lvert mt_i \rvert$: size of a multi-test.*<br>$Cost(T_i) = \frac{\lvert X \rvert}{\lvert X_i \rvert} * C(a_{ij})$<br>*where: $a_{ij}$: j-th attribute of the i-th multi-test, C($a_{ij}$): cost of the $a_{ij}$ attribute.* | • Used for fitting decision trees.<br>• This function needs to be maximized to achieve high performance. |
| 39 | F2 score [112] | Measured as the weighted average of precision and recall. | $F_\beta = 1 + \beta 2 \times \frac{precision \times recall}{(\beta 2 \times precision) + recall}$<br>*where: β = 2.* | • Used in genetic programming and medical fields.<br>• Computes a weighted harmonic mean of Precision and Recall.<br>• Learning about the minority class. |
| 40 | Distance score (D score) [112] | - | $D_{sc} = \frac{2 \times C1 \times C2}{C1 + C2}$<br>*where:*<br>$C1 = \frac{\sum_{i=0}^{N_{maj}} sig(P_{Maji}) \times \lvert T - sig(P_{Maji}) \rvert}{N_{maj}} \times func(1, P_{Maji})$<br>$sig(x) = \frac{2}{1 + e - x} - 1$<br>$C2 = \frac{\sum_{i=0}^{N_{min}} sig(P_{Mini}) \times \lvert T - sig(P_{Mini}) \rvert}{N_{min}} \times func(1, P_{Mini})$<br>$func(1,k) = \begin{cases} 1, & \text{if} k \leq 0 \text{formajorityclassinstanc} \\ 1, & \text{if} k > 0 \text{forminorityclassinstanc} \\ 0, & \text{otherwise} \end{cases}$<br>*C1* for majority class and *C2* for minority class. | • Used in genetic programming and medical fields.<br>• Distance score (D score) which learns about both the classes by giving them equal importance and being unbiased.<br>• The range of both C1 and C2 is 0 (worst score) to 1 (best score). |

*The reader is encouraged to review the cited references for full details on specific metrics.

## 3. Closing Remarks

Our confidence in the accuracy of predictions obtained from ML algorithms heavily relies on the availability of actual observations and proper PFEMs. From this point of view, it is unfortunate that observations relating to the engineering discipline continue to be 1) limited in size, and 2) lack completeness. The lack of such observations is often related to limitations in conducting full scale tests, need for specialized equipment, and wide variety in tested samples. For instance, one can think of how normal strength concrete mixes can significantly vary from one study to another simply due to variation in raw materials, mix proportions and casting/curing procedure, etc.

Combining the above two points with the notion of simply "applying ML" to understand a given phenomenon (say flexural strength of beams) without a thorough validation is deemed to fail. In fact, in many instances, researchers noted the validity of a specific ML model by reporting its performance against traditional PFEMs, only to be later identified that such a model does not properly represent actual observations – despite having good fitness. This can be avoided by adopting a rigorous validation procedure [113,114]. Unfortunately, many of the published studies in the area of ML application in engineering do not include multi-criteria/additional validation phases and simply rely on conventional performance metrics such as $R$ or $R^2$ of the derived models. Furthermore, adopting a set of PFEMs does not negate the occurrence of some common issues most notably, overfitting, biasedness etc. As such, an analysis that utilizes ML should also considers some of the following techniques e.g. use of independent test datasets, varying degrees of cross-validation etc.

In order to ensure fruitful use of ML, it is our duty to seek proper application of ML. Besides, one of the major concerns about the ML-based models their robustness under a wide range of conditions [115]. A robust ML model should not only provide reasonable PFEMs but should also be capable of capturing the underlying physical mechanisms that govern the investigated system [116] . An essential approach to verify the robustness of the ML models is to perform parametric and sensitivity analyses [115,117]. These types of analyses ensure that the ML predictions are in a sound agreement with the system's real behavior and physical processes rather than being merely a combination of the variables with the best fit on the data. Another item to consider is to develop a user-friendly phenomenon-specific recommendation system wherein novice users apply pre-identified PFEMs are selected to evaluate the performance of a given problem (say using $R^2$ in a regression problem etc.).

The reader is to remember that the addition of one example to showcase recommended or important PFEMs negates the purpose of this paper. It is our intention to not specifically identify a measure (or a set of measures) due to the wide range of problems (as well as quality of data) that a scientist could face. Please note that other researchers (which are quoted herein) also followed a similar approach.

- *"Although some methods clearly perform better or worse than other methods on average, there is significant variability across the problems and metrics. Even the best models sometimes perform poorly, and models with poor average performance occasionally perform exceptionally well."* [118].
- *"It is clearly difficult to convincingly differentiate ML algorithms (and feature reduction techniques) on the basis of their achievable accuracy, recall and precision."*[119].
- *"Different performance metrics yield different tradeoffs that are appropriate in different settings. No one metric does it all, and the metric optimized to or used for model selection does matter."*[94].

## 4. Conclusions

Based on the information presented in this note, the following conclusions can be drawn.

- ML is expected to rise into a key analysis tool in the coming few years; especially within material scientists and structural engineers.
- The integration of ML is to be thorough and proper. Hence, the need for proper validation procedure.
- A variety of performance metrics and error metrics exists for regression and classification problems. This work recommends the utilization of multi-fitness criteria to ensure validity of ML models as these metrics may overcome some of the limitations of induvial metrics.
- The performance of the existing metrics as well as future fitness functions can be further improved through a systematic collaboration between researchers of interdisciplinary backgrounds.
- Future works should be directed towards documenting and exploring performance metrics for other types of learnings such as unsupervised learning, reinforcement learning. This is an ongoing research need that is to be addressed in the coming years.